\documentclass[conference]{IEEEtran}
\IEEEoverridecommandlockouts
\usepackage{cite}
\usepackage{amsmath,amssymb,amsfonts}
\usepackage{algorithmic}
\usepackage{graphicx}
\usepackage{textcomp}
\usepackage{xcolor}
\usepackage{booktabs}
\usepackage{makecell}
\usepackage{hyperref}

\def\BibTeX{{\rm B\kern-.05em{\sc i\kern-.025em b}\kern-.08em
    T\kern-.1667em\lower.7ex\hbox{E}\kern-.125emX}}
\begin{document}

\title{Prompt-Based Caption Generation for Single-Tooth Dental Images Using Vision-Language Models\\

% {\footnotesize \textsuperscript{*}Note: Sub-titles are not captured for https://ieeexplore.ieee.org  and
% should not be used}
}

\author{\IEEEauthorblockN{Anastasiia Sukhanova}
  \IEEEauthorblockA{\textit{Marshall University} \\
    asukhanova@marshall.edu}
  \and
  \IEEEauthorblockN{Aiden Taylor}
  \IEEEauthorblockA{\textit{Marshall University} \\
    taylor1144@marshall.edu}
  \and
  \IEEEauthorblockN{Julian Myers}
  \IEEEauthorblockA{\textit{Marshall University} \\
    myers254@marshall.edu}
  \and
  \IEEEauthorblockN{Zichun Wang}
  \IEEEauthorblockA{\textit{West Virginia State University} \\
    zwang2@wvstateu.edu}
  \and
  \IEEEauthorblockN{Kartha Veerya Jammuladinne}
  \IEEEauthorblockA{\textit{West Virginia State University} \\
    kjammuladinne@wvstateu.edu}
  \and
  \IEEEauthorblockN{Satya Sri Rajiteswari Nimmagadda}
  \IEEEauthorblockA{\textit{Marshall University} \\
    nimmagadda2@marshall.edu}
  \and
  \IEEEauthorblockN{Aniruddha Maiti}
  \IEEEauthorblockA{\textit{West Virginia State University} \\
    aniruddha.maiti@wvstateu.edu}
  \and
  \IEEEauthorblockN{Ananya Jana}
  \IEEEauthorblockA{Marshall University\\
    jana@marshall.edu}

}

\maketitle
\begin{abstract}
Digital dentistry has made significant advances with the advent of deep learning. However, the majority of these deep learning-based dental image analysis models focus on very specific tasks such as tooth segmentation, tooth detection, cavity detection, and gingivitis classification. There is a lack of a specialized model that has holistic knowledge of teeth and can perform dental image analysis tasks based on that knowledge. Datasets of dental images with captions can help build such a model. To the best of our knowledge, existing dental image datasets with captions are few in number and limited in scope. In many of these datasets, the captions describe the entire mouth, while the images are limited to the anterior view. As a result, posterior teeth such as molars are not clearly visible, limiting the usefulness of the captions for training vision-language models. Additionally, the captions focus only on a specific disease (gingivitis) and do not provide a holistic assessment of each tooth. Moreover, tooth disease scores are typically assigned to individual teeth, and each tooth is treated as a separate entity in orthodontic procedures. Therefore, it is important to have captions for single-tooth images. As far as we know, no such dataset of single-tooth images with dental captions exists. In this work, we aim to bridge that gap by assessing the possibility of generating captions for dental images using Vision-Language Models (VLMs) and evaluating the extent and quality of those captions. Our findings suggest that guided prompts help VLMs generate meaningful captions. We show that the prompts generated by our framework are better anchored in describing the visual aspects of dental images. We selected RGB images as they have greater potential in consumer scenarios.
\end{abstract}

\begin{figure}[htbp]
  \centering
  \includegraphics[width=0.49\textwidth]{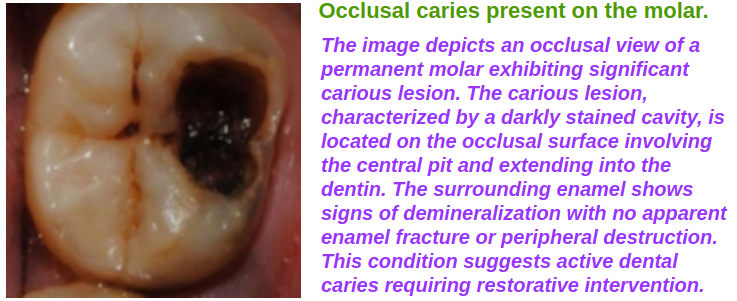}
  \caption{\small Example of captions generated using Vision Language Model GPT-4o. These captions include tooth surface, number, and condition. The short caption is written in green and the long caption is written in violet.}
  \label{fig:captioning_colgate}
\end{figure}

\section*{Introduction}
Deep learning methods have significantly impacted dentistry by automating tasks such as tooth segmentation \cite{wang2025tooth,jana20233d, chen2025cross, jing2025teethnet, xi20253d,lee2025render2seg
}, tooth detection \cite{beser2024yolo, ghorbani2025novel, ozccelik2025optimized}, dental crown generation \cite{chen2026precise, hosseinimanesh2025personalized}, and synthetic tooth data generation \cite{jana2024development}. Though these models are good at their individual tasks in isolation, there is a lack of a specialized vision-language model that possesses a holistic understanding of any dental diagnostic task. In general computer vision, we have models such as CLIP \cite{radford2021learning} and BLIP \cite{li2022blip}, and in the medical imaging domain, we have models such as MedCLIP \cite{wang2022medclip} and BioCLIP \cite{stevens2024bioclip}, which are trained for domain-specific tasks.
\begin{figure*}[t]
\centering
\includegraphics[width=0.9\textwidth]{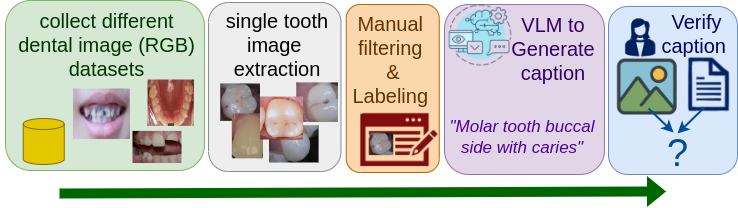}
\caption{\small Publicly available dental images are collected and categorized by view. They are processed with a tooth detector, filtered, captioned using prompt-guided vision-language models aligned with dental ontologies, and finally evaluated.}
\label{fig:pipeline}
\end{figure*}
These models do not generate proper captions for dental images mainly because they do not have an understanding of the dental domain (e.g., MedCLIP is trained on X-ray images and BLIP is trained on natural images). The dental vision-language model DenVLM \cite{meng2025dentvlm} is trained on a large dental dataset, but it is also based on full-mouth images. The OralGPT \cite{hao2025towards} model is trained on dental panoramic X-ray images, which are different from RGB images. Currently, there is no dental image-based vision-language model that works on single-tooth images.  Training such a specialized model for single-tooth images would require a large number of dental image-caption pairs. It is in this context that we investigate the efficacy of vision-language models (namely, GPT-4o) in generating dental image captions directly from unannotated dental images.Our framework produces structured, clinically aligned captions for individual intraoral tooth images, incorporating tooth number, surface, and condition using dental ontologies. 

Public datasets of intraoral dental images are commonly used to train models for classification or detection. These datasets usually contain labels for specific diagnostic categories such as caries, staining, or white lesions. They do not include structured descriptions of tooth anatomy or multi-attribute captions. Labels are often limited to one condition per image. As a result, existing datasets do not support tasks that require structured image-to-text generation or holistic representation of dental findings. Natural language descriptions offer a compact and interpretable way to summarize multiple attributes per image. A caption that combines tooth number, surface, and condition reflects common clinical language and improves reuse across tasks. Dental disease scores are usually given to individual teeth rather than the entire jaw or mouth. Thus, to build a specialized model with a holistic understanding of the tooth, we will need individual tooth images with captions describing their conditions. However, the only dental image caption dataset \cite{duy2024dental} has captions for the entire mouth when images are only from the anterior view. Moreover, the focus is only on the disease gingivitis and its location rather than a visual description of the disease. Such a caption may not sufficiently capture the visual aspects of the entire image. Moreover, the caption provides gingivitis scores for teeth that are not adequately visible in the front view or anterior image. Another drawback is that the captions are not natural language-style captions.

Our proposed framework generates captions aimed at overcoming these challenges. There are publicly available dental RGB images, but often these RGB images come in the form of full-mouth images. However, the availability of individual dental images is crucial since dental disease scoring is done for individual teeth. The available datasets for dental images suffer from various problems such as labels available for only specific diseases, low-quality images (poor lighting \cite{chaudhary2024teeth}, blurring \cite{dental-dataset-87ne6-aitkd_dataset}, artifacts, motion artifacts, pixelation), occlusion, varying resolutions \cite{dentalmate_dataset, mouthdetection_dataset}, lack of surface information (e.g., lingual, occlusal, or buccal), limited views (e.g., only anterior views or a mix of anterior and occlusal views), augmented images mixed with original data \cite{dentalmate_dataset}, data duplicacy \cite{chaudhary2024teeth}, and inconsistency in the number of teeth per image, e.g., some have only a single tooth while others have multiple teeth in focus.

This paper presents a framework for generating a structured captions for single tooth dental images, illustrated in Figure~\ref{fig:pipeline}. The framework described consists of collecting data from different publicly available repositories, segregating the images into different categories according to the different views, excluding problematic images by manual evaluation, generating single-tooth images from images with multiple teeth with a tooth detection algorithm, labeling them in separate categories and excluding the problematic images again, and designing a prompt to generate single tooth image captions in a two-step prompt engineering process.

The main contributions of this paper are as follows:
\begin{itemize}
  \item We propose a framework to generate captions for dental images without requiring pre-existing annotations, using a two-step prompt engineering strategy with vision-language models.
  \item We curate and process a diverse collection of publicly available RGB intraoral images, extracting high-quality single-tooth views across multiple dental surfaces and conditions.
  \item We demonstrate that structured prompts improve caption quality by guiding the VLM to include clinically relevant details such as tooth type, surface, and disease condition.
  \item We evaluate the accuracy of the generated captions through both automated label comparisons and manual review.
\end{itemize}

\begin{figure*}[t]
\centering
\includegraphics[width=\textwidth]{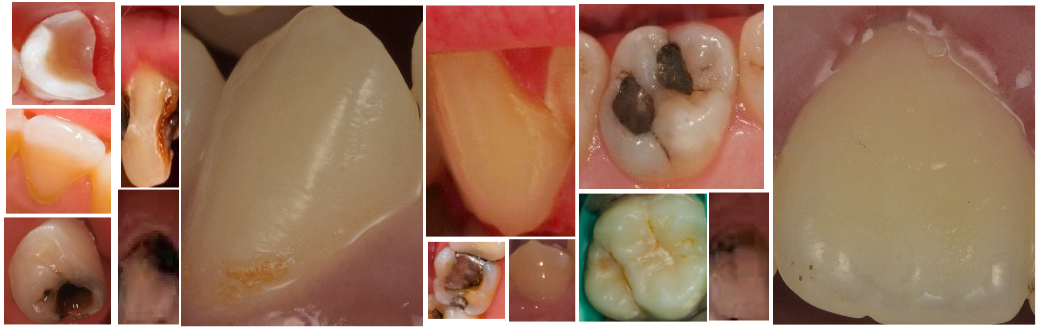}
\caption{\small Examples of diverse tooth images curated from publicly available repositories.}
\label{fig:teethexamples}
\end{figure*}

\section*{Method}
\subsection*{Dataset}
We collected dental images from four publicly available datasets — two of the datasets from Mendeley and the remaining two datasets from Roboflow. We curated images having diverse views. The first dataset we used was the Mendeley gingivitis captioning dataset \cite{duy2024dental} whcih contained anterior views. The Dental or Intraoral Teeth image dataset \cite{chaudhary2024teeth} from Mendeley contains images of children's teeth between the ages of 1 and 14. This dataset includes multiple views of teeth. The mouth detection dataset \cite{mouthdetection_dataset} from Roboflow had different anterior view images, and they contained images where the teeth had different diseases and captured images of teeth from various age groups, such as kids to grown-ups. The DentalMate \cite{dental-dataset-87ne6-aitkd_dataset} from Roboflow contained many images with occlusal views, primarily of the premolar and molar teeth, and different tooth conditions such as caries, tooth loss, tooth decay, etc. 

\subsection*{Dataset Preparation and Preprocessing}
Dataset preparation began with the selection of image datasets that contained visible teeth. This step was necessary to make the data suitable for later procedures. Each dataset went through a cleaning process. We excluded images with staining, poor image quality, or other visual problems that could result in invalid captions. We used a neural network trained for tooth detection. This network generated single-tooth image samples from each image in the dataset. These generated images were filtered again. We removed pictures with low resolution or poor quality to keep the dataset consistent and usable. We used three types of views: occlusal view, anterior view, and buccal view. The occlusal view shows the teeth from directly above. It captures the chewing surface of molars and premolars clearly. The anterior view shows the front teeth at the front of the mouth. The buccal view shows the outer surfaces of teeth that face the cheek. These are mostly molars and premolars. We curated a total of 6 datasets curated from the 4 original datasets as described.
%From the gingivitis captioning dataset\cite{duy2024dental} from Mendeley we selected only the anterior views of incisor and canine images, as these are most prominently visible. From  Dental or Intraoral Teeth image dataset \cite{chaudhary2024teeth} from Mendeley, we selected three view types. The first subset includes buccal views of canine and premolar teeth from the Lower Right and Lower Left views and molar teeth from the Lower Right and Lower Left views. The third subset includes occlusal views of premolar and molar teeth taken from the lower occlusal views.In this manner, we have total of 6 datasets curated from those 4 original datasets.

\textbf{Dataset 1} (Mendeley gingivitis dataset~\cite{duy2024dental}) included buccal or anterior images.\\
\textbf{Dataset 2} (Roboflow mouth detection dataset~\cite{mouthdetection_dataset}) contained buccal or anterior views.\\
\textbf{Dataset 3} (Mendeley dataset~\cite{chaudhary2024teeth}) provided buccal views from the upper right and upper left jaws.\\
\textbf{Dataset 4} (Mendeley dataset~\cite{chaudhary2024teeth}) included occlusal surfaces of premolar and molar teeth.\\
\textbf{Dataset 5} (Mendeley dataset~\cite{chaudhary2024teeth}) consisted of posterior buccal views of premolar and molar teeth.\\
\textbf{Dataset 6} (Roboflow DentalMate dataset~\cite{dental-dataset-87ne6-aitkd_dataset}) mainly included occlusal views of molar and premolar teeth.

The distribution of image counts per tooth type across these datasets is shown in Table~\ref{tab:simple_tooth_dataset_final}.

\begin{figure}
  \centering
  \includegraphics[width=0.45\textwidth]{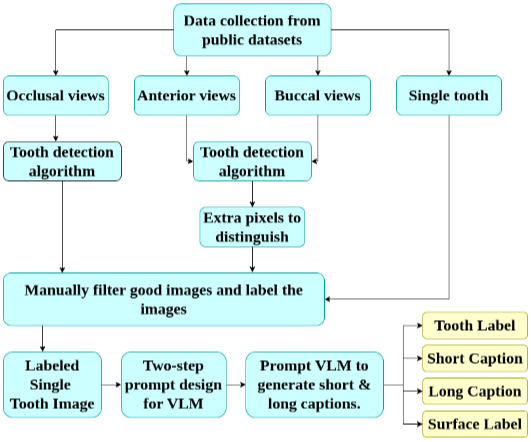}

\caption{\small Detailed flowchart of the proposed framework. This diagram outlines each step in the pipeline — from dataset collection and filtering, to tooth detection, prompt-based caption generation, and evaluation. The two-step prompt engineering strategy is a key component of the caption generation process.}
  \label{fig:flowchart}
\end{figure}

A visual summary of the full preprocessing pipeline, including dataset selection, filtering, and single-tooth extraction, is shown in Figure~\ref{fig:flowchart}.

\begin{table}[ht]
  \centering
  \caption{Summary of single tooth datasets with total image counts per dataset.}
  \label{tab:simple_tooth_dataset_final}
  \begin{tabular}{|l|r|r|r|r|r|}
    \hline
    \textbf{Dataset} & \textbf{Incisor} & \textbf{Canine} & \textbf{Premolar} & \textbf{Molar} & \textbf{Total} \\ \hline
    Dataset 1 & 250  & 250  & -   & -   & 500  \\ \hline
    Dataset 2 & 411  & 107  & -   & -   & 518  \\ \hline
    Dataset 3 & -    & 238  & 236 & -   & 474  \\ \hline
    Dataset 4 & -    & -    & 43  & 44  & 87   \\ \hline
    Dataset 5 & -    & -    & 65  & 47  & 112  \\ \hline
    Dataset 6 & -    & -    & -   & -   & 498  \\ \hline
  \end{tabular}
\end{table}

\begin{figure*}[t]
  \centering
  \includegraphics[width=.4\textwidth]{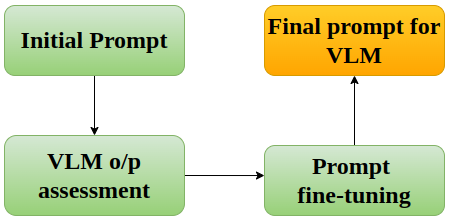}
  \includegraphics[width=.45\textwidth]{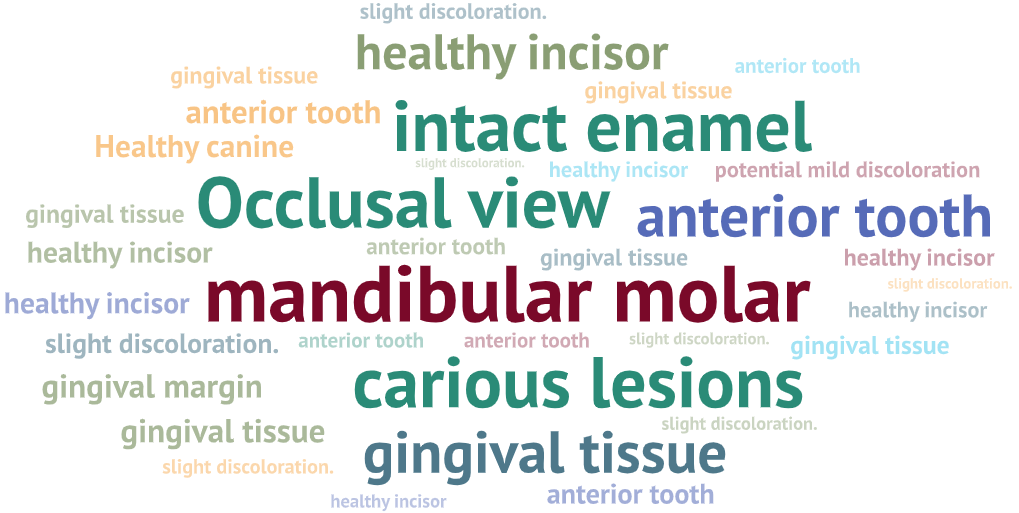}
  \caption{\small Illustration of our two-step prompt engineering strategy (left), including basic and refined prompts, and a word cloud showing the most frequent clinical keywords extracted from the generated captions (right).}
  \label{fig:caption}
\end{figure*}

We cleaned all images manually based on the earlier filtering criteria. We did not apply the tooth detection algorithm to the already single-tooth images in Dataset 6. We applied the detection algorithm to the images containing multiple teeth to generate single-tooth images on the Datasets numbered 1 to 5. After detection, we cropped the images. For buccal views and anterior views, we added 60 extra pixels to include the gum region of each tooth. This gave more context about tooth condition. We did not add pixels to occlusal views because gums are already hidden in those images.

We manually filtered the cropped images and kept only those with clear tooth structure. We used full-mouth or partial-mouth references to label the tooth type. Experts annotated 433 of 500 selected single-tooth images from Dataset 6 by identifying tooth surfaces. We did not categorize the remaining images because the teeth were too damaged. From anterior views, we kept incisor and canine teeth. Other types were not fully visible. We grouped central incisors and lateral incisors under the single label “incisor.”

\subsection*{Framework}
As the next step, we used a Vsion Language Model (VLM) GPT-4o to generate both short and long captions for each tooth to test its descriptive capabilities. This step tested the VLM’s ability to recognize and describe various tooth and gum conditions. Based on the captions generated in this first step, we analyzed the results to refine the next prompt, which produced the final captions. The second prompt also evaluated image quality and marked low-quality images. After the second prompt, we filtered 1,520 good images out of 2,308 images (which were initially cleaned) based on quality. The captions associated with these good images were then analyzed.

\subsection*{Experiments}
We originally had 2,308 images. We processed the images using online processing with the maximum number of retries set to 5. We used a ChatGPT Plus account, which ran GPT-4o to generate the captions. The batch processing job was sent to GPT-4o in online processing mode.

\subsection*{Prompt Engineering}
Since vision-language models may extract information from filenames and directories, we applied a preprocessing step to avoid unintentional label leakage. In many publicly available intraoral image datasets, the file or folder names contain clinically relevant terms such as disease types (e.g., “caries”, “fracture”), anatomical identifiers (e.g., “tooth32”), or positional labels. Exposing such labels to the captioning model could result in unintended bias or label leakage during generation, especially if the model uses filename text through internal mechanisms such as OCR or metadata parsing. To mitigate this, we recursively scanned all images in the input directory and copied them to a temporary working directory with anonymized filenames. Each image was renamed to a generic format (img\_0001.jpg, img\_0002.jpg, etc.), and a mapping between the original and anonymized paths was maintained. Only the anonymized images were used during batch caption generation.

The first-level prompt asked the VLM to generate basic short and long captions for each dental image.

We observed that the model often labeled teeth as anterior rather than distinguishing between incisors and canines. After the initial captions were generated, we manually inspected a small subset of outputs (about 20 samples per dataset split) to identify recurring weaknesses. These included misclassification of anterior teeth, incomplete surface descriptions, and omission of visual disease cues.
\begin{figure*}[t]
  \centering
  \includegraphics[width=\textwidth]{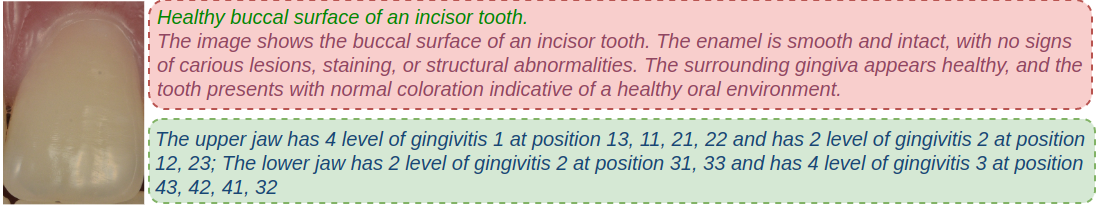}
  \caption{\small Comparison of captions from the gingivitis caption dataset~\cite{duy2024dental} (red box) and VLM-generated captions (green box). Please note that the gingivitis dataset caption describes the gingivitis condition from the whole-mouth image from which this single-tooth image was taken.}
  \label{fig:captcompare}
\end{figure*}

Based on these observations, we reformulated the prompts into more explicit and context-aware instructions.  The second-level prompt included more detailed instructions to extract specific labels such as tooth type, surface, and disease before generating both types of captions. An overview of the two-step prompt design and the corresponding keywords extracted from the generated captions (from the first level prompt)is shown in Figure~\ref{fig:caption}.  The two-step process led the model to generate captions that were both clinically grounded and semantically consistent. The refinement process improved caption reliability, especially in distinguishing anterior and posterior teeth and in identifying dental pathologies such as caries and discoloration.

\subsection*{Evaluation}
As discussed earlier, in the second stage of prompting, the VLM evaluated image quality (e.g., blurry or poor-resolution images). Based on this, we retained only the high-quality images identified by the model, resulting in a total of 1,520 images with the tooth type distribution shown in Table~\ref{tab:tooth_dataset_summary}.

The filtered dataset was then used to assess caption quality.  
We evaluated the quality of the generated captions using two methods:  
(i) automatic evaluation, which measured the accuracy of tooth number and surface labels produced by the VLM;  
(ii) manual evaluation, where we reviewed a small set of captions from each category to estimate overall descriptive accuracy.

Both evaluations used accuracy as the performance metric. Here, accuracy refers to the percentage of images where the VLM's predicted tooth type, surface, or condition matched the expert label or ground truth. The automatic evaluation measured the correctness of tooth number and surface labels, while the manual evaluation assessed how often the captions correctly described actual tooth conditions, as verified by the expert.

\begin{table}
  \centering
  \caption{Summary of single tooth datasets with total image counts per category (after filtering good quality images).}
  \label{tab:tooth_dataset_summary}
  \begin{tabular}{|l|l|l|l|l|r|}
    \hline
    Dataset & incisor & canine & premolar & molar & Total \\
    \hline
    Dataset 1 & 250 & 244 & - & - & 494 \\
    \hline
    Dataset 2 & 346 & 84 & - & - & 430 \\
    \hline
    Dataset 3 & - & 88 & 69 & - & 157 \\
    \hline
    Dataset 4 & - & - & 3 & 9 & 12 \\
    \hline
    Dataset 5 & - & - & 9 & 4 & 13 \\
    \hline
    Dataset 6 & 4 & 6 & 90 & 258 & 358 \\
    \hline
  \end{tabular}
\end{table}
\section*{Results}
\subsection*{Performance in labeling tooth type and surface type}
We evaluated the performance of the final prompt, which directed the VLM to assign both tooth type and surface labels to each image. In many cases, the VLM made better decisions regarding tooth type while generating short and long captions, as shown in Table~\ref{tab:combined_accuracy}. The second column reports the accuracy of the tooth type assigned directly by the VLM, while the third column shows the accuracy of the tooth type inferred from its captions. The fourth column presents the model’s performance in identifying the tooth surface.

\begin{figure*}
  \centering
  \includegraphics[width=\textwidth]{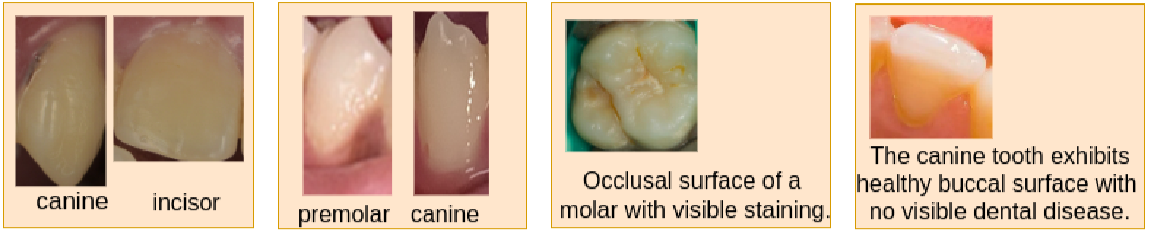}
  \caption{\small Examples of tooth\_types that confused the VLMs and images where the VLM produced an incorrect caption.}
  \label{fig:teethexamples1}
\end{figure*}

As seen in Table~\ref{tab:combined_accuracy}, the VLM occasionally misclassifies adjacent tooth types, such as confusing canines with incisors or premolars, and molars with premolars. In addition to label accuracy, we compared the generated captions with existing expert-labeled captions from public datasets, as illustrated in Figure~\ref{fig:captcompare}. To better understand the types of errors made by the model, we examined several examples where the VLM produced incorrect captions or misclassified tooth types, as shown in Figure~\ref{fig:teethexamples1}.

\begin{table}
  \centering
  \caption{Combined accuracy metrics across datasets with record counts. The tooth\_type is the label assigned by the VLM, whereas inferred\_type is the tooth type inferred from the short and long captions.}
  \label{tab:combined_accuracy}
  \begin{tabular}{l|rrr}
    \toprule
    Dataset & tooth\_type & inferred\_type & surface \\
    \midrule
   Dataset 1 (494) & 0.5526      & 0.6296         & 0.8846  \\
     Dataset 2 (430) & 0.2116      & 0.7488         & 0.3070  \\
    Dataset 3 (157) & 0.0892      & 0.0892         & 0.8471  \\
    Dataset 4 (12)  & 0.7500      & 0.7500         & 1.0000  \\
    Dataset 5 (13)  & 0.2308      & 0.1538         & 0.9231  \\
    Dataset 6 (358) & 0.7067      & 0.7263         & 0.9638  \\
    \bottomrule
  \end{tabular}
\end{table}

\subsection*{Visual Assessment}
We performed a visual evaluation of caption quality using a random sample from each dataset split. An expert examined captions for common dental conditions such as caries, staining, enamel loss, and discoloration.

The expert estimated how accurately the captions described specific conditions. Table~\ref{tab:disease_summary} shows the percentage of correct assessments for selected disease categories.

\begin{table}
  \centering
  \caption{Disease accuracy summary table showing the percentage of correct assessments for selected diseases.}
  \label{tab:disease_summary}
  \begin{tabular}{l|rrrr}
    \toprule
    Dataset  & caries & staining & enamel & discoloration \\
    \midrule
    Dataset 1 (21) & 95.24  & 95.24    & 80.95  & 90.48          \\
    Dataset 2 (20) & 100.00 & 100.00   & 95.00  & 95.00          \\
    Dataset 3 (17) & 100.00 & 82.35    & 82.35  & 94.12          \\
    Dataset 4 (11) & 72.73  & 90.91    & 100.00 & 90.91          \\
    Dataset 5 (13) & 100.00 & 100.00   & 92.31  & 92.31          \\
    Dataset 6 (20) & 70.00  & 80.00    & 80.00  & 95.00          \\
    \bottomrule
  \end{tabular}
\end{table}

Our dataset did not include images of the lingual surface. This limited our ability to evaluate caption accuracy for that region. In some cases, light reflecting off the tooth surface was misinterpreted by the model as demineralization.

In Dataset 1, canine teeth were captured from anterior views. As a result, the VLM often labeled them as anterior teeth instead of canines. In such views, the canine appears triangular, resembling an incisor because part of the tooth is out of focus or not visible.

In Dataset 3, the model initially labeled many incisor teeth as molars. However, it corrected this after considering the full caption content. In the second prompt, we did not include any gum-specific diseases such as bleeding. We observed that the model struggled with detecting gingivitis. In the first round of captioning, it often labeled images from the gingivitis dataset as having healthy gums, which was inaccurate. The difficulty likely stems from the subtle visual changes caused by gingivitis compared to more noticeable conditions like caries or demineralization. Although extra padding was added to retain gum regions in anterior images, it may not have been enough for the model to assess gingival health properly.

Nonetheless, the model showed strong ability in recognizing more visible features such as tooth type, surface, caries, structural damage, wear, discoloration, and demineralization. In the second prompt, we guided it to evaluate these conditions and any others it could detect.

In addition to challenges with disease detection, we observed confusion in surface-type classification. The VLM often misclassified the buccal surface of canines as premolars and the occlusal surface of molars as premolars. It also confused canines with incisors in anterior views, likely due to partial visibility and triangular shape. Shadows along the middle ridge of canines may have contributed to this confusion. The model also struggled with deciduous (child) teeth, which differ in shape and proportion from adult teeth.

Overall, the visual assessment showed that the VLM can describe major structural and disease-related features with reasonable accuracy. However, it struggles with subtle distinctions such as gingival conditions, surface differentiation, and partial views. These findings highlight the need for better prompt design and more comprehensive datasets to improve diagnostic precision.

We can question the necessity of a specialized model for the dental domain, given that GPT-4o was able to generate captions with reasonable success. However, since we cannot train or fine-tune GPT-4o for specific dental tasks, we are limited by its current capabilities. Thus, creating a framework for caption generation is an important step toward building a model with specialized understanding of all types of dental diagnostic tasks.

\subsection*{Limitations}

While our framework shows promising results, several limitations remain. First, the vision-language model (GPT-4o) is not fine-tunable, which restricts domain-specific adaptation. Second, our dataset lacks lingual surface views, limiting the completeness of surface-level evaluation. Third, the model struggled to identify subtle clinical signs such as early-stage gingivitis or gum inflammation, possibly due to insufficient gum visibility or due to the subtle nature of the disease. Fourth, deciduous (child) teeth with irregular morphology led to higher misclassification rates, especially when occlusion was present. Additionally, although we used expert validation on a subset, full-scale expert review was not done across the entire dataset. Finally, the reliance on publicly available RGB images constrained the diversity and consistency of dental conditions represented.

These limitations highlight the need for improved dataset coverage, and eventually, the development of fine-tunable dental-specific vision-language models.

\subsection*{Discussion}
The results presented above demonstrate the potential of using a vision-language model to generate clinically meaningful captions for dental images. This work introduces a framework for generating captions for publicly available dental images without using ground truth annotations. The currently available tooth captioning dataset\cite{duy2024dental} does not provide any visual guidance except the location and severity of gingivitis. An example caption is: \textit{"The upper jaw has teeth 12 and level of gingivitis 3."} Compared to this caption, our captions offer much better visual guidance for understanding tooth structure. 

In practice, a single tooth can exhibit multiple conditions simultaneously, requiring a comprehensive understanding for accurate assessment. Such a captioned dataset helps move toward building a specialized model for dental images analysis. Clinicians often assess gum health when evaluating overall tooth condition. To capture this, we had expanded the bounding boxes of front teeth by 60 pixels to include the gum at the root. However this adjustment did not seem to help in assessing the gingivitis disease.

In some cases, the VLM labeled images as good, but the corresponding short descriptions indicated poor quality. The model was initially prompted to assess image quality before generating captions. However, its quality evaluation appeared to improve during caption generation, possibly due to deeper visual analysis at that stage. A similar pattern was observed in tooth type labeling, where the assigned label improved after the model processed the image more fully during caption creation.

\section*{Conclusion}

This work presents a framework for generating structured, clinically meaningful captions for dental images using vision-language models and prompt engineering. By curating a diverse set of publicly available RGB intraoral images and applying a two-step prompting strategy, we produce captions that include tooth type, surface, and condition without requiring manual annotation. Evaluation shows that guided prompts significantly improve caption quality, supporting the feasibility of building annotated datasets from uncaptioned images. This framework lays the groundwork for developing specialized vision-language models tailored to dental diagnostics.

\section*{Acknowledgments}
The image captions were generated using GPT-4o.
\noindent Part of the computational requirements for this work were supported by resources at \textit{Delta at NCSA} and \textit{Anvil at Purdue University} through \textbf{ACCESS allocation Grant number: CIS250068 and CIS250766} from the \textit{Advanced Cyberinfrastructure Coordination Ecosystem: Services and Support (ACCESS)} program, which is supported by U.S. National Science Foundation grants \#2138259, \#2138286, \#2138307, \#2137603, and \#2138296.

\section*{Compliance with Ethical Standards}
\label{sec:ethics}
This research study was conducted retrospectively using human subject data made available in open-access repositories by Mendeley and Roboflow. Ethical approval was not required, as confirmed by the licenses associated with the open-access datasets.

\bibliographystyle{IEEEbib}
\bibliography{refs}

\end{document}